\documentclass[runningheads]{llncs}
\usepackage{graphicx}
\usepackage{amsmath,amssymb} 
\usepackage{color}
\usepackage{tabularx}
\usepackage{booktabs}
\usepackage{colortbl}
\usepackage{multirow}
\usepackage[table]{xcolor}
\usepackage{slashbox}
\usepackage{extarrows}
\usepackage{hyperref}
\usepackage{enumerate}
\usepackage{subfig}
\usepackage{bm}
\usepackage{rotating}
\usepackage{color}
\usepackage{cite}
\usepackage[ruled,vlined]{algorithm2e}
\definecolor{shadecolor}{rgb}{0.95,0.95,0.95}
\newcommand{\mycolor}[1]{\multicolumn{1}{>{\columncolor{shadecolor}}c}{{}#1}}
\renewcommand{\paragraph}{\textbf}
\makeatletter
\newcommand{\printfnsymbol}[1]{%
	\textsuperscript{\@fnsymbol{#1}}%
}
\renewcommand*{\@fnsymbol}[1]{\ensuremath{\ifcase#1\or *\or \dagger\or \ddagger\or \mathsection\or \mathparagraph\or \|\or **\or \dagger\dagger \or \ddagger\ddagger \else\@ctrerr\fi}}
\makeatother

\begin{document}

\title{Highly-Economized Multi-View Binary Compression for Scalable Image Clustering}

\titlerunning{Highly-Economized Multi-View Binary Compression}

\author{Zheng Zhang\inst{1,2,3}\thanks{ indicates equal contributions; $\dagger$ indicates the corresponding author.} \and Li Liu\inst{3}\printfnsymbol{1} \and Jie Qin\inst{4}\printfnsymbol{1} \and Fan Zhu\inst{3} \and Fumin Shen\inst{5} \and Yong Xu\inst{1}\printfnsymbol{2} \and Ling Shao\inst{3} \and Heng Tao Shen\inst{5}}

\authorrunning{Z. Zhang et al.}

\institute{Harbin Institute of Technology (Shenzhen), China \and
The University of Queensland, Australia \and
Inception Institute of Artificial Intelligence, UAE \and
Computer Vision Laboratory, ETH Zurich, Switzerland \and
University of Electronic Science and Technology of China, China}

\maketitle

\begin{abstract}
How to economically cluster large-scale multi-view images is a long-standing problem in computer vision. To tackle this challenge, we introduce a novel approach named \textbf{Highly-economized Scalable Image Clustering (HSIC)} that radically surpasses conventional image clustering methods via binary compression. We intuitively unify the binary representation learning and efficient binary cluster structure learning into a joint framework. In particular, common binary representations are learned by exploiting both sharable and individual information across multiple views to capture their underlying correlations. Meanwhile, cluster assignment with robust binary centroids is also performed via effective discrete optimization under $\ell_{21}$-norm constraint. By this means, heavy continuous-valued Euclidean distance computations can be successfully reduced by efficient binary XOR operations during the clustering procedure. To our best knowledge, HSIC is the first binary clustering work specifically designed for scalable multi-view image clustering. Extensive experimental results on four large-scale image datasets show that HSIC consistently outperforms the state-of-the-art approaches, whilst significantly reducing \textit{computational time} and \textit{memory footprint}.
\end{abstract}

\section{Introduction}
Image clustering is a commonly used unsupervised analytical technique for practical computer vision applications \cite{AKJ2010}. The aim of image clustering is to discover the natural and interpretable structure of image representations, so as to group images that are similar to each other into the same cluster. Based on the number of sources where images are collected or number of features how images are described, existing clustering methods can be divided into single-view image clustering (SVIC) \cite{kmeans,sculley2010web,otto2018clustering,avrithis2015web} and multi-view\footnote{\scriptsize Despite `multi-view' can refer to multiple features, domains or modalities, in this paper, we solely focus on the clustering problem for images with multiple features (\textit{e.g.}, LBP, HOG and GIST).}  image clustering (MVIC) \cite{bickel2004,xu2013survey,xia2010multiview,MultiNMF,MVKM}. Recently, MVIC \cite{bickel2004,xu2013survey,zhang2017marginal} has been evoking more and more attention due to the flexibility of extracting multiple heterogeneous features from a single image. Compared to SVIC, MVIC has access to more characteristics and structural information of the data, and the features from diverse views can potentially complement each other and produce more effective clustering performance.

Existing MVIC methods can be roughly divided into three groups: multi-view spectral clustering \cite{kumar2011co,AMGL,MLAN}, multi-view matrix factorization \cite{MultiNMF,MVKM,OMVC}, and multi-view subspace clustering \cite{GaoMV,wang2017exclusivity,zhang2017latent}. Multi-view spectral clustering \cite{xia2010multiview} constructs multiple similarity graphs to achieve a common or similar eigenvector matrix on all views, and then generates consensus data partitions, which hinge crucially on the single-view spectral clustering \cite{ng2002spectral}. Due to the straightforward interpretability of matrix factorization \cite{NMF}, multi-view matrix factorization methods \cite{MultiNMF,MVKM} integrate information from multiple views towards a compatible common consensus, or decompose the heterogeneous features into specified centroid and cluster indicator matrices. Different from the above strategies, multi-view subspace clustering \cite{GaoMV} employs the complementary properties across multiple views to uncover the common latent subspace and quantify the genuine similarities. Some other kernel-based MVIC methods \cite{tzortzis2012kernel,de2010multi} exploit a linear or a non-linear kernel on each view. Note that SVIC (\textit{e.g.}, $k$-means \cite{kmeans} and spectral clustering \cite{ng2002spectral}) can also be leveraged to deal with multi-view clustering problem. A common practice for them is to perform clustering on either any single-view feature or simply concatenated multiple features \cite{xu2013survey,xia2010multiview}.

\begin{figure}[!t]
\centering\resizebox{1\textwidth}{!}{
\includegraphics[width=7in]{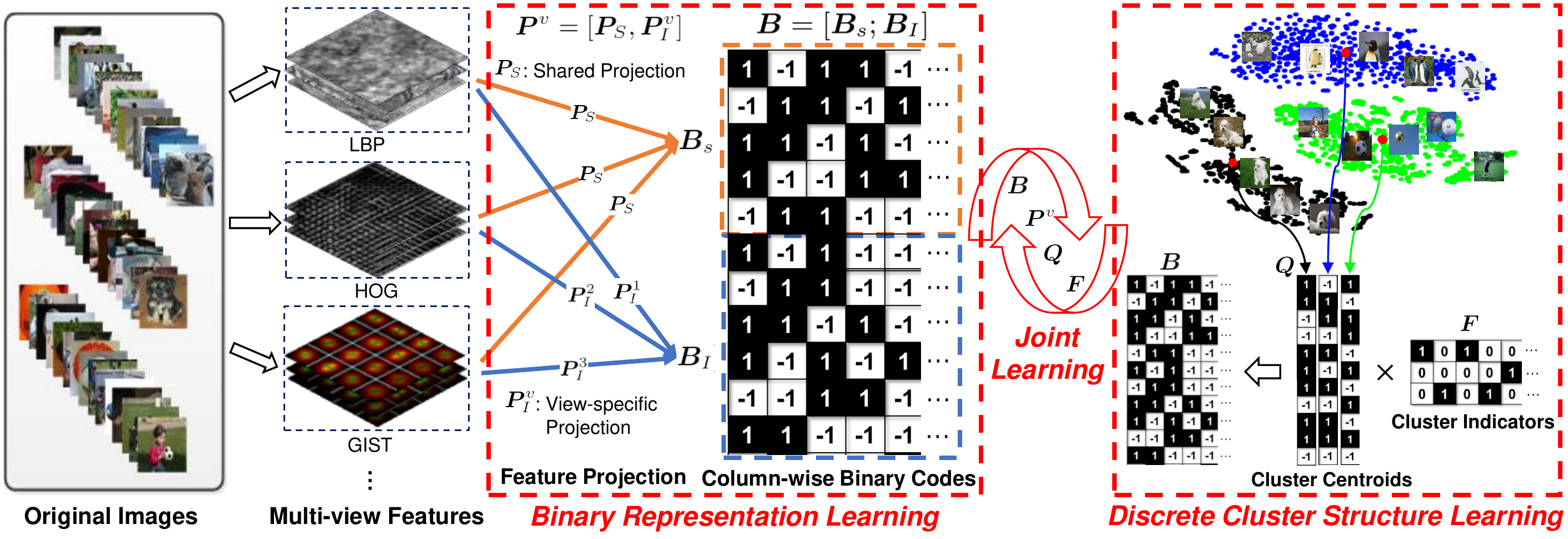}}
\caption{\small The pipeline of HSIC. Common binary representation learning and discrete cluster structure learning are jointly and efficiently solved by alternating optimization.}\label{fig_1}
\end{figure}

Although SVIC and MVIC methods have achieved much progress on small- and middle-scale data, both of them will become intractable (because of unaffordable computation and memory overhead) when dealing with large-scale data with high dimensionality, which is a typical case in the era of `big data'. As pointed out in \cite{gong2015web,shen2017compressed}, we argue that real-valued features are the essential bottleneck restricting the scalability of existing clustering methods. To address this issue, inspired by the recent advances on compact binary coding (\emph{a.k.a.} hashing) \cite{wang2017survey,shen2015cvpr,shen2017sigir,liulatent2017,liu2016sequential,qin2017cvpr1,chen2017cvpr,lu2017simultaneous}, we aim to develop a feasible binary clustering technique for large-scale MVIC. Specifically, we transform the original real-valued Euclidean space to the low-dimensional binary Hamming space, based on which an efficient clustering solution can then be devised. In this way, time-consuming Euclidean distance measures (typically of $\mathcal{O}(Nd)$ complexity, where $N$ and $d$ respectively indicate the data size and dimension) for real-valued data can be substantially eliminated by the extremely fast XOR operations (of $\mathcal{O}$(1) complexity) for compact binary codes. Note that the proposed method is also potentially promising in practical use cases where computation and memory resources are limited (\textit{e.g.}, on wearable or mobile devices).

As shown in Fig. \ref{fig_1}, we particularly develop a \textit{Highly-economized Scalable Image Clustering} (HSIC) framework for efficient large-scale MVIC. HSIC jointly learns the effective common binary representations and robust discrete cluster structures. The former can maximally preserve both sharable and view-specific/individual information across multiple views; the latter can significantly promote the computational efficiency and robustness of clustering. The joint learning strategy is superior to separately learning each objective by facilitating the collaboration between both objectives. An efficient alternating optimization algorithm is developed to address the joint discrete optimization problem. The main contributions of this work include:

1) To the best of our knowledge, HSIC is the pioneering work with large-scale MVIC capability, where common binary representations and robust binary cluster structures can be obtained in a unified learning framework.

2) HSIC captures both sharable and view-specific information from multiple views to fully exploit the complementation and individuality of heterogeneous image features. The sparsity-induced $\ell_{21}$-norm is imposed on the clustering model to further alleviate its sensitivity against outliers and noise.

3) Extensive experimental results on four image datasets clearly show that HSIC can reduce the \textit{memory footprint} and \textit{computational time} up to \textbf{951} and \textbf{69.35} times respectively over the classical \textit{k}-means algorithm, whilst consistently outperform the state-of-the-art approaches.

Notably, two works \cite{gong2015web,shen2017compressed} in the literature are most relevant to ours. \cite{gong2015web} introduced a two-step binary $k$-means approach, in which clustering is performed on the binary codes obtained by Iterative Quantization (ITQ) \cite{gong2013iterative}, and \cite{shen2017compressed} integrated binary structural SVM and $k$-means. Our HSIC fundamentally differs from them in the following aspects: 1) \cite{gong2015web} and \cite{shen2017compressed} are SVIC methods, while HSIC is specially designed for MVIC; 2) \cite{gong2015web} divides the clustering task into two unconnected procedures, which completely eliminate the important tie between the binary coding and cluster structure learning. Meanwhile, the binary codes learned by \cite{shen2017compressed} are too weak to achieve satisfactory results because of lacking adequate representative capability. More importantly, both methods cannot make full use of the complementary properties of multiple views for scalable MVIC, which is also shown in \cite{BMVC}.

In the next section, we will introduce the detailed framework of our HSIC and then elaborate on the alternating optimization algorithm. The analysis in terms of computational complexity and memory load will also be presented.

\normalsize
\section{Highly-economized Scalable Image Clustering}
Suppose we have a set of multi-view image features $\mathcal X$ = $\{\bm X^1,\cdots,\bm X^m\}$ from $m$ views, where $\bm X^v = [\bm x_1^v, \cdots, \bm x_N^v] \in \Re^{d_v \times N}$ is the accumulated feature matrix from the $v$-th view. $d_v$ and $N$ denote the dimensionality and the number of data points in $\bm X^v$, respectively. $\bm x_i^v \in \Re^{d_v \times 1}$ is the $i$-th feature vector from the $v$-th view. The main objective of unsupervised MVIC is to partition $\mathcal X$ into $c$ groups, where $c$ is the number of clusters. In this work, to address the large-scale MVIC problem, our HSIC aims to perform binary clustering in the much lower-dimensional Hamming space. Particularly, we perform multi-view compression (\textit{i.e.}, project multi-view features onto the common Hamming space) by learning the compatible \textbf{common binary representation} via the complimentary characteristics of multiple views. Meanwhile, \textbf{robust binary cluster structures} are formulated in the learned Hamming space for efficient clustering.

As a preprocessing step, we first normalize the features from each view as zero-centered vectors. 	Inspired by \cite{shen2015cvpr,liu2011hashing}, in this work, each feature vector is encoded by the simple nonlinear RBF kernel mapping, \textit{i.e.}, $\psi(\bm x_i^v) = [exp(-\|\bm x_i^v - \bm a_1^v\|^2/\gamma),\cdots, exp(-\|\bm x_i^v - \bm a_l^v\|^2/\gamma)]^{\top}$,
where $\gamma$ is the pre-defined kernel width, and $\psi(\bm x_i^v)\in \Re^{l\times 1}$ denotes an $l$-dimensional nonlinear embedding for the $i$-th feature from the $v$-th view. Similar to \cite{shen2015cvpr,liu2011hashing,liu2014discrete}, $\{\bm a_i^v\}_{i=1}^l$ are randomly selected $l$ anchor points from $\bm X^v$  ($l=1000$ is used for each view in this work). Subsequently, we will introduce how to learn the common binary representation and robust binary cluster structure respectively, and finally end up with a joint learning objective.

\noindent\textbf{1) Common Binary Representation Learning}. We consider a family of $K$ hashing functions to be learned in HSIC, which quantize each $\psi(\bm x_i^v)$ into a binary representation $\bm b_i^v = [b_{i1}^v,\cdots,b_{iK}^v]^T \in \{-1,1\}^{K\times 1}$. To eliminate the semantic gaps between different views, HSIC generates the common binary representation by combining multi-view features. Specifically, HSIC simultaneously projects features from multiple views onto a common Hamming space, \textit{i.e.,} $\bm b_i = sgn\big((\bm P^v)^{\top} \psi(\bm x_i^v)\big)$,
where $\bm b_i$ is the common binary code of the $i$-th features from different views (\textit{i.e.,} $\bm x_i^v$, $\forall v=1,...,m$), $sgn(\cdot)$ is an element-wise sign function, $ \bm P^v = [\bm p_1^v, \cdots, \bm p_K^v] \in \Re^{l\times K}$ is the mapping matrix for the $v$-th view and $\bm p_i^v$ is the projection vector for the $i$-th hashing function. As such, we construct the learning function by minimizing the following quantization loss:
\begin{equation} \label{eq_3}\small
\min_{\bm P^v,\bm b_i} \sum_{v=1}^m \sum_{i=1}^N \| \bm b_i - (\bm P^v)^{\top} \psi(\bm x_i^v)\|_F^2.
\end{equation}
Since different views describe the same subject from different perspectives, the projection $\{\bm P^v\}_{v=1}^m$ should capture the shared information that maximizes the similarities of multiple views, as well as the view-specific/individual information that distinguishes individual characteristics between different views. To this end, we decompose each projection into the combination of sharable and individual projections, \textit{i.e.,}  $\bm P^v = [\bm P_S, \bm P_I^v]$. Specifically, $\bm P_S \in \Re^{l\times K_S}$ is the shared projection across multiple views, while $\bm P_I^v \in \Re^{l\times K_I}$ is the individual projection for the $v$-th view, where $K = K_S+K_I$. Therefore, HSIC collectively learns the common binary representation from multiple views using
\small\begin{align} \label{eq_4}
&\min_{\bm P^v, \bm B,\alpha^v} \sum_{v=1}^m (\alpha^v)^r \big(\| \bm B - (\bm P^v)^{\top} \psi(\bm X^v)\|_F^2+\lambda_1 \|\bm P^v\|_F^2\big),\nonumber\\
s.t.~&\sum_v \bm\alpha^v=1,\bm\alpha^v>0, \bm B = [\bm B_s; \bm B_I]\in \{-1,1\}^{K \times N}, \bm P^v = [\bm P_s, \bm P_I^v],
\end{align}
where $\bm B = [\bm b_1,\cdots,\bm b_N]$, $\bm \alpha =[\alpha^1,\cdots,\alpha^m]\in \Re^m$ weighs the importance of different views, $r>1$ is a constant managing the weight distributions, and $\lambda_1$ is a regularization parameter. The second term is a regularizer to control the parameter scales.

Moreover, from the information-theoretic point of view, the information
provided by each bit of the binary codes needs to be maximized \cite{Baluja2008}. Based on this point and motivated by \cite{gong2013iterative,wang2010semi}, we adopt an additional regularizer for the binary codes $\bm B$ using maximum entropy
principle, \textit{i.e.}, $\max~var[\bm B] = var[sgn\big((\bm P^v)^{\top} \psi(\bm x_i^v)\big)]$. This additional regularization on $\bm B$ can ensure the \textit{balanced partition} and \textit{reduce the redundancy} of the binary codes. Here we replace the sign function by its signed magnitude, and formulate the relaxed regularization as follows
\small\begin{align} \label{eq_5}
\max \sum_k \mathbb{E}[\|(\bm p_i^v)^{\top} \psi(\bm x_i^v)\|^2] = \frac{1}{N} tr\big((\bm P^v)^{\top} \psi(\bm X^v) \psi(\bm X^v)^{\top}\bm P^v\big) = g(\bm P^v).
\end{align}

Finally, we combine problems (\ref{eq_4}) and (\ref{eq_5}) together and reformulate the overall common binary representation learning problem as the following
\small\begin{align} \label{eq_6}
& \min_{\bm P^v, \bm B} \sum_{v=1}^m (\alpha^v)^r \big(\| \bm B - (\bm P^v)^{\top} \psi(\bm X^v)\|_F^2 + \lambda_1 \|\bm P^v\|_F^2 - \lambda_2 g(\bm P^v)\big)\nonumber\\ &s.t.~ \sum_v \alpha^v=1,\alpha^v>0,\bm B = [\bm B_s; \bm B_I] \in \{-1,1\}^{K \times N}, \bm P^v = [\bm P_s, \bm P_I^v],
\end{align}
where $\lambda_2$ is a weighting parameter.

\noindent\textbf{2) Robust Binary Cluster Structure Learning}. For binary clustering, HSIC directly factorizes the learned binary representation $\bm B$ into the binary clustering centroids $\bm Q$ and discrete clustering indicators $\bm F$ using
\begin{align} \label{eq_7}\small
\min_{\bm Q, \bm F} \| \bm B - \bm Q\bm F\|_{21},~s.t.~\bm Q\bm 1 = \bm 0,\bm Q \in \{-1,1\}^{K\times c},\bm F \in \{0,1\}^{c\times N}, \sum_j f_{ji} = 1,
\end{align}
where $\|\bm A\|_{21} = \sum_i\|\bm a^i\|_2$, and $\bm a^i$ is the $i$-th row of matrix $\bm A$. The first constraint of (\ref{eq_7}) ensures the balanced property on the clustering centroids as with the binary codes. Note that the $\ell_{21}$-norm imposed on the loss function can also be replaced by the $F$-norm, \textit{i.e.}, $\| \bm B - \bm Q\bm F\|_F^2$. However, the $F$-norm based loss function can amplify the errors induced from noise and outliers. Therefore, to achieve more stable and robust clustering performance, we employ the sparsity-induced $\ell_{21}$-norm. It is also observed in \cite{ding2006r} that the $\ell_{21}$-norm not only preserves the rotation invariance within each feature, but also controls the reconstruction error, which significantly mitigates the negative influence of the representation outliers.

\noindent\textbf{3) Joint Objective Function}. To preserve the semantic interconnection between the learned binary codes and the robust cluster structures, we incorporate the common binary representation learning and the discrete cluster structure constructing into a joint learning framework. In this way, the unified framework can interactively enhance the qualities of the learned binary representation and cluster structures. Hence, we have the following joint objective function:
\small\begin{align} \label{eq_9}
& \min_{\bm P^v,\bm B,\bm Q,\bm F,\alpha^v} \sum_{v=1}^m (\alpha^v)^r \big(\| \bm B - (\bm P^v)^{\top} \psi(\bm X^v)\|_F^2 + \lambda_1 \|\bm P^v\|_F^2 - \lambda_2 g(\bm P^v)\big)+\lambda_3\| \bm B - \bm Q\bm F\|_{21},\nonumber\\ &s.t.~\sum_v \bm\alpha^v=1,\bm\alpha^v>0,\bm B = [\bm B_s; \bm B_I] \in \{-1,1\}^{K \times N}, \bm P^v = [\bm P_s, \bm P_I^v],\nonumber\\ &~~~~~~\bm Q\bm 1 = \bm 0, \bm Q \in \{-1,1\}^{K\times c},\bm F \in \{0,1\}^{c\times N}, \sum_j f_{ji} = 1,
\end{align}
where $\lambda_1$, $\lambda_2$ and $\lambda_3$ are trade-off parameters to balance the effects of different terms. To optimize the difficult discrete programming problem, a newly-derived alternating optimization algorithm is developed as shown in the next section.

\begin{algorithm}[!t]
  \caption{Highly-economized Scalable Image Clustering (HSIC)}
  \label{alg_1}
  \SetKwInOut{Input}{Input}
  \SetKwInOut{Output}{Output}
  \SetKwInOut{Init}{Initial.}
  \Input{Multi-view features $\{\bm{X}^v\}_{v=1}^m \in \Re^{d_v\times N}$, $m\geq 3$; code length $K$; number of centroids $c$; maximum iterations $\kappa$ and $t$; $\lambda_1$, $\lambda_2$ and $\lambda_3$.}
  \Output{Binary representation $\bm B$, cluster centroid $\bm Q$ and cluster indicator $\bm F$.}
\Init{Randomly select $l$ anchor points from each view to calculate the kernelized feature embedding $\psi(\bm X^v)\in \Re^{l\times N}$, and normalize them to have zero-centered mean.}
\Repeat{convergence or reach $t$ iterations}{
\textbf{$\bm P_S$-Step}: Update $\bm P_S$ by Eqn.(\ref{eq_12});\\
\textbf{$\bm P_I^v$-Step}: Update $\bm P_I^v$ by Eqn. (\ref{eq_13}), $\forall v=1,\cdots,m$;\\
\textbf{$\bm B$-Step}: Update $\bm B$ by Eqn. (\ref{eq_18});\\
\Repeat{convergence or reach $\kappa$ iterations}{
\textbf{$\bm Q$-Step:} Iteratively update $\bm Q$ by Eqn. (\ref{eq_21});\\
\textbf{$\bm F$-Step:} Update $\bm F$ by Eqn. (\ref{eq_25});\\
}
\textbf{$\bm\alpha$-Step}: Update $\bm\alpha$ by Eqn. (\ref{eq_28});\\
}
\end{algorithm}

\subsection{Optimization}
The solution to problem (\ref{eq_9}) is non-trivial as it involves a mixed binary integer program with three discrete constraints, which lead to an NP-hard problem. In the following, we introduce an alternating optimization algorithm to iteratively update each variable while fixing others, \textit{i.e.}, update $\bm{P}_s\rightarrow \bm P_I^v \rightarrow \bm{B} \rightarrow \bm{Q} \rightarrow \bm{F} \rightarrow \bm \alpha$ in each iteration.

Due to the intractable $\ell_{21}$-norm loss function, we first rewrite the last term in (\ref{eq_9}) as $\lambda_3 tr\big(\bm U^{\top} \bm D\bm U\big)$, where $\bm U = \bm B - \bm Q\bm F$, and $\bm D \in \Re^{K\times K}$ is a diagonal matrix, the $i$-th diagonal element of which is defined as $\bm d_{ii}$ = $1/ 2\|\bm u^i\|$, where $\bm u^i$ is the $i$-th row of $\bm U$.

\noindent\textbf{1) $\bm{P}_s$-Step}: When fixing other variables, we update the sharable projection by
\small\begin{align} \label{eq_11}
&\min_{\bm P_s} \sum_{v=1}^m (\alpha^v)^r \big( \| \bm B_s - \bm P_s^{\top} \psi(\bm X^v)\|_F^2 + \lambda_1 \| \bm P_s\|_F^2 - \frac{\lambda_2}{N} tr\big(\bm P_s^{\top} \psi(\bm X^v) \psi^{\top}(\bm X^v)\bm P_s\big)\big).
\end{align}
For notational convenience, we rewrite $\psi(\bm X^v)\psi^{\top}(\bm X^v)$ as $\tilde{\bm X}$. Taking derivation of $\mathcal L$ with respect to $\bm P_s$ and let $\frac{\partial \mathcal L}{\partial \bm P_s} = 0$, we can obtain the closed-form solution of $\bm P_s$, \textit{i.e.,}
\small\begin{align} \label{eq_12}
\bm P_s = (\bm A+\lambda_1 \sum_{v=1}^m (\alpha^v)^r\bm I)^{-1} \bm T \bm B^{\top},
\end{align}
where $\bm A = (1-\frac{\lambda_2}{N})\sum_{v=1}^m (\alpha^v)^r \tilde{\bm X}$ and $\bm T = \sum_{v=1}^m (\alpha^v)^r \psi(\bm X^v)$.

\noindent\textbf{2) $\bm P_I^v$-Step}: Similarly, when fixing other parameters, the optimal solution of the $v$-th individual projection matrix can be determined by solving
\small\begin{align} \label{eq_13}
&\min_{\bm P_I^v} \| \bm B_I - (\bm P_I^v)^{\top} \psi(\bm X^v)\|_F^2 + \lambda_1 \|\bm P_I^v\|_F^2 - \frac{\lambda_2}{N} tr\big(\bm P_I^v  \tilde{\bm X}(\bm P_I^v)^{\top}\big),
\end{align}
and its closed-form solution can be obtained by $\bm P_I^v = \bm W\psi(\bm X^v)\bm{B}^{\top}$, where $\bm W = \left((1-\frac{\lambda_2}{N})\tilde{\bm X}+\lambda_1 \bm I\right)^{-1}$ can be calculated beforehand.

\noindent\textbf{3) $\bm{B}$-Step}: Problem (\ref{eq_9}) w.r.t. $\bm B$ can be rewritten as:
\small\begin{align} \label{eq_15}
\min_{\bm B} \sum_{v=1}^m (\alpha^v)^r \big(\| \bm B - (\bm P^v)^{\top} \psi(\bm X^v)\|_F^2 \big)+\lambda_3 tr\big(\bm U^{\top} \bm D\bm U \big),
~s.t.~\bm B \in \{-1,1\}^{K \times N}.
\end{align}
Since $\bm B$ only has `1' and `-1' entries and $\bm D$ is a diagonal matrix, both $tr(\bm B \bm B^{\top})$ = $tr(\bm B^{\top}\bm B) = KN$ and $tr\left(\bm{B}^{\top}\bm{DB}\right)$ = $N*tr(\bm{D})$ are constant terms w.r.t. $\bm B$. Based on this and with some further algebraic computations, (\ref{eq_15}) can be reformulated as
\begin{align} \label{eq_17}
\resizebox{0.92\linewidth}{!}{$
\min\limits_{\bm B} - 2 tr\left[ \bm B^{\top} \left(\sum_{v=1}^{m}(\alpha^v)^r \big((\bm{P}^v)^{\top}\psi(\bm X^v)\big)+\lambda_3 \bm{QF}\right)\right]+const,s.t.~\bm{B}\in \{-1,1\}^{K\times N},
$}
\end{align}
where `\textit{const}' denotes the constant terms. This problem has a closed-form solution:
\small\begin{align} \label{eq_18}
\bm B =sgn \left(\sum_{v=1}^{m}(\bm\alpha^v)^r \big((\bm{P}^v)^{\top}\psi(\bm X^v)\big) + \lambda_3 \bm{QF}\right).
\end{align}

\noindent\textbf{4) $\bm{Q}$-Step}: First, we degenerate (\ref{eq_9}) into the following computationally feasible problem (by removing some irrelevant parameters and discarding the first constraint):
\small\begin{align}\label{eq_20}
\min_{\bm Q, \bm F}tr\big(\bm U^{\top} \bm D\bm U \big)+\nu\|\bm{Q}^{\top}\bm 1\|_F^2,~s.t.~\bm{Q}\in \{-1,1\}^{K\times c},~\bm F \in \{0,1\}^{c\times N}, \sum_j f_{ji} = 1.
\end{align}
With sufficiently large $\nu> 0$, problems (\ref{eq_9}) and (\ref{eq_20}) will be equivalent. Then, by fixing the variable $\bm{F}$, problem (\ref{eq_20}) becomes
\small\begin{align}\label{eq_21}
\min_{\bm Q} \mathcal L(\bm Q) = -2tr(\bm B^{\top} \bm D \bm{QF})+\nu \|\bm{Q}^{\top}\bm 1\|_F^2 + const,~s.t.~\bm{Q}\in \{-1,1\}^{K\times c}.
\end{align}
Inspired by the efficient discrete optimization algorithm in \cite{TIP2016binary,qin2017cvpr}, we develop an adaptive discrete proximal linearized optimization algorithm, which iteratively updates $\bm Q$ in the ($p$+$1$)-th iteration by $\bm Q^{p+1} = sgn(\bm Q^p-\frac{1}{\eta}\nabla \mathcal L(\bm Q^p))$,
where $\nabla\mathcal L(\bm Q)$ is the gradient of $\mathcal L(\bm Q)$, $\frac{1}{\eta}$ is the learning step size and $\eta \in(C,2C)$, where $C$ is the Lipschitz constant. Intuitively, for the very special $sgn(\cdot)$ function, if the step size $1/\eta$ is too small/large, the solution of $\bm Q$ will get stuck in a bad local minimum or diverge. To this end, a proper $\eta$ is adaptively determined by enlarging or reducing based on the changing values of $\mathcal L(\bm Q)$ between adjacent iterations, which can accelerate its convergence.

\noindent\textbf{5) $\bm{F}$-Step}: Similarly, when fixing $\bm Q$, the problem w.r.t. $\bm F$ turns into
\small\begin{align}\label{eq_23}
\min_{\bm f_i} \sum_{i=1}^N \bm d_{ii}\| \bm b_i - \bm Q\bm f_i\|_{21},~s.t.~\bm f_i \in \{0,1\}^{c\times 1}, \sum_j f_{ji} = 1.
\end{align}
We can divide the above problem into $N$ subproblems, and independently optimize the cluster indicator in a column-wise fashion. That is, one column of $\bm F$ (\textit{i.e.}, $\bm f_i$) is computed at each time. Specifically, we solve the subproblems in an exhaustive search manner, similar to the conventional $k$-means algorithm.
Regarding the $i$-th column $\bm f_i$, the optimal solution of its $j$-th entry can be efficiently obtained by
\small\begin{equation} \label{eq_25}
f_{ji} = \left\{
\begin{aligned}
1&,&j = \arg\min_k H(\bm d_{ii}*\bm b_i,\bm q_{\wp}),
\\0&,&otherwise,
\end{aligned} \right.
\end{equation}
where $\bm q_{\wp}$ is the $\wp$-th vector of $\bm Q$, and $H(\cdot,\cdot)$ denotes the Hamming distance metric. Note that computing the Hamming distance is remarkably faster than the Euclidean distance, so the assigned vector $\bm f_i$ will efficiently constitute the matrix $\bm F$.

\noindent\textbf{6) $\bm \alpha$-Step}: Let $\displaystyle h^v$ = $\|\bm{B}-\left(\bm{P}^v\right)^{\top}\phi(\bm X^v)\|_F^2 + \lambda_1 \| \bm{P}^v ||_F^2 -\lambda_2 g(\bm{P}^v)$, then problem (\ref{eq_9}) w.r.t. $\bm\alpha$ can be rewritten as
\small\begin{align}\label{eq_26}
\min_{\alpha^v} \sum_{v=1}^{m} (\alpha^v)^r h^v,~s.t.~\sum_v \alpha^v=1,\alpha^v>0.
\end{align}
The Lagrange function of (\ref{eq_26}) is $\min \mathcal L(\alpha^v,\bm\zeta) = \sum_{v=1}^{m} (\alpha^v)^r h^v-\bm\zeta (\sum_{v=1}^{m} \alpha^v-1)$,
where $\bm\zeta$ is the Lagrange multiplier. Taking the partial derivatives w.r.t. $\alpha^v$ and $\bm\zeta$, respectively, we can get
\small\begin{equation} \label{eq_28}
\left\{
\begin{aligned}
\frac{\partial \mathcal L}{\partial\alpha^v} &=& r(\alpha^v)^{r-1}h^v-\bm\zeta,\\
\frac{\partial \mathcal L}{\partial\bm\zeta} &=& \sum_{v=1}^{m} \alpha^v-1.
\end{aligned}\right.
\end{equation}
Following \cite{xia2010multiview}, by setting $\nabla_{\alpha^v,\bm\zeta} \mathcal L$ = $\bm 0$, the optimal solution of {\small$\alpha^v$} is $\frac{(h^v)^{\frac{1}{1-r}}}{\sum_v(h^v)^{\frac{1}{1-r}}}$.

To obtain the locally optimal solution of problem (\ref{eq_9}), we update the above six variables iteratively until convergence. To deal with the out-of-example problem in image clustering, HSIC needs to generate the binary code for a new query image $\hat{\bm x}$ from the $v$-th view (\textit{i.e.}, $\hat{ \bm x}^v$) by ${\bm b}^{v} = sgn\left((\bm{P}^v)^{\top}\psi(\hat{\bm x}^v)\right)$, and then assigns it to the $j$-th cluster decided by $j = \arg\min_k H({\bm b}^{v},\bm q_k)$ in the fast Hamming space. For multi-view clustering, the common binary code of $\hat{\bm x}$ is $\bm b =sgn \left(\sum_{v=1}^{m}(\bm\alpha^v)^r (\bm{P}^v)^{\top}\psi(\hat{\bm x}^v)\right)$. Then the optimal cluster assignment of $\hat{\bm x}$ is determined by the solution of $\bm F$. The full learning procedure of HSIC is illustrated in Algorithm \ref{alg_1}.

\subsection{Complexity and Memory Load Analysis}
\textbf{1)} The major computation burden of HSIC lies in the compressive binary representation learning and robust discrete cluster structures learning. The computational complexities of calculating $\bm P_S$ and $\bm P_I^v$ are $\mathcal{O}(K_SlN)$ and $\mathcal{O}(m(K_IlN))$, respectively. Computing $\bm B$ consumes $\mathcal{O}(KlN)$. Similar to \cite{gong2015web}, constructing the discrete cluster structures needs $\mathcal{O}(N)$ on bit-wise operators for $\kappa$ iterations, where the distance computation requires only $\mathcal{O}(1)$ per time. The total computational complexity of HSIC is $\mathcal{O}(t((K_S+mK_I+K)lN+\kappa N))$, where $t$ and $\kappa$ are empirically set to 10 in all the experiments. In general, the computational complexity of optimizing HSIC is linear to the number of samples, \textit{i.e.}, $\mathcal{O}(N)$. \textbf{2)} For memory cost in HSIC, it is unavoidable to store the mapping matrices $\bm P_s$ and $\bm P_I^v$, demanding $\mathcal{O}(lK_S)$ and  $\mathcal{O}(lK_I)$ memory costs, respectively. Notably, the learned binary representation and discrete cluster centroids only need the \textbf{\textit{bit-wise}} memory load $\mathcal{O}(K(N+c))$, which is much less than that of $k$-means requiring $\mathcal{O}(d(N+c))$ \textbf{\textit{real-valued}} numerical storage footprint.

\section{Experimental Evaluation}\label{Exp}
In this section, we conducted multi-view image clustering experiments on four scalable image datasets to evaluate the effectiveness of HSIC with four frequently-used performance measures. All the experiments are implemented based on Matlab 2013a using a standard Windows PC with an Intel 3.4 GHz CPU.

\subsection{Experimental Settings}\label{Exset}
\textbf{Datasets and Features:} We perform experiments on four image datasets, including \textbf{ILSVRC2012 1K} \cite{deng2009imagenet}, \textbf{Cifar-10} \cite{krizhevsky2009learning}, \textbf{YouTube Faces} (YTBF) \cite{wolf2011face} and \textbf{NUS-WIDE} \cite{NUSWIDE}. Specifically, we randomly select $10$ classes from ILSVRC2012 1K with $1,300$ images per class, denoted as \textit{ImageNet-10}, for \textit{middle-scale} multi-view clustering study. \textit{Cifar-10} contains $60,000$ tiny color images in $10$ classes, with $6,000$ images per class. A subset of \textit{YTBF} contains $182,881$ face images from $89$ different people ($>1,200$ for each one). Similar to \cite{TIP2016binary}, we collect the subset of \textit{NUS-WIDE} including the $21$ most frequent concepts, resulting in $195,834$ images with at least $3,091$ images per category. Because some images in NUS-WIDE were labeled by multiple concepts, we only select one of the most representative labels as their true categories for simplicity. Multiple features are extracted on all datasets. Specifically, for ImageNet-10, Cifar-10 and YTBF, we use three different types of features, \textit{i.e.}, $1450$-d LBP, $1024$-d GIST, and $1152$-d HOG. For NUS-WIDE, five publicly available features are employed for experiments, \textit{i.e.}, $64$-d color Histogram (CH), $225$-d color moments (CM), $144$-d color correlation (CORR), $73$-d edge distribution (EDH) and $128$-d wavelet texture (WT).

\textbf{Metrics and Parameters:} We adopt four widely-used evaluation measures \cite{manning2008} for clustering, including clustering accuracy (ACC), normalized mutual information (NMI), purity, and F-score. In addition, both computational time and memory footprint are compared to show the efficiency of HSIC. To fairly compare different methods, we run the provided codes with default or fine-tuned parameter settings according to the original papers. For binary clustering methods, 128-bit code length is used for all datasets. For hyper-parameters $\lambda_1$, $\frac{\lambda_2}{N}$, and $\lambda_3$ of HSIC, we first employ the grid search strategy on ImageNet-10 to find the best values (\textit{i.e.}, $10^{-3}$, $10^{-3}$, and $10^{-5}$, respectively), which are then directly adopted on other datasets for simplicity. We empirically set $r$ and $\delta = \frac{K_S}{K}$ (\textit{i.e.}, the ratio of shared binary codes) as $5$ and $0.2$ respectively in all experiments. The multi-view clustering results are denoted as `MulView'. We report the averaged clustering results with $10$ times randomly initialization for each method.

We conduct the following experiments from \textbf{\textit{three}} perspectives. \textbf{\textit{Firstly}}, we verify various characteristics of HSIC on the \textit{middle-scale} dataset, \textit{i.e.}, ImageNet-10. Here we compare HSIC with both SVIC and MVIC methods (including real-valued and binary ones). \textbf{\textit{Secondly}}, three large-scale datasets are exploited to evaluate HSIC on the challenging large-scale MVIC problem. \textbf{Remark:} Based on the results on ImageNet-10 (see Table \ref{Table_2}), the real-valued MVIC methods only obtain comparable results to $k$-means, but they are very time-consuming. Moreover, when applying those MVIC methods (\textit{e.g.,} AMGL and MLAN) to larger datasets, we encounter the `out-of-memory' error. Therefore, the real-valued MVIC methods are not compared on the three large-scale datasets. \textbf{\textit{Thirdly}}, some empirical analyses of our HSIC are also provided.

\subsection{Experiments on the Middle-Scale ImageNet-10}
We compare HSIC to several state-of-the-art clustering methods including SVIC methods (\textit{i.e.,} $k$-means \cite{kmeans}, $k$-Medoids\cite{park2009simple}, Approximate kernel $k$-means\cite{AKKmeans}, Nystr\"{o}m \cite{chen2011parallel}, NMF \cite{NMF}, LSC-K \cite{chen2011large}), MVIC methods, (\textit{i.e.,} AMGL \cite{AMGL}, MVKM \cite{MVKM}, MLAN \cite{MLAN}, MultiNMF \cite{MultiNMF}, OMVC \cite{OMVC}, MVSC \cite{li2015large}) and two existing binary clustering methods (\textit{i.e.,} ITQ+$bk$-means \cite{gong2015web} and CKM \cite{shen2017compressed}). Additionally, two variants of HSIC are also compared to show its efficacy, \textit{i.e.,} HSIC with $F$-norm regularized binary clustering (HSIC-F), and HSIC with two separate steps of binary code learning and discrete clustering (HSIC-TS). Similar to \cite{MultiNMF,li2015large}, for all the SVIC methods, we simply concatenate the feature vectors of all views for the `MulView' clustering.

\begin{table*}[!t]
\begin{center}
\caption{\small Performance comparisons on ImageNet-10. The bold \textbf{black} and {\color{blue}{\textbf{blue}}} numbers indicate the best single-view and multi-view clustering results, respectively.}
\label{Table_1}
\resizebox{0.99\textwidth}{!}{
\begin{tabular}{c|lccccccccccccccccccccc}
\toprule
\multicolumn{2}{c}{Metric}  && \multicolumn{4}{c}{ACC} && \multicolumn{4}{c}{NMI} && \multicolumn{4}{c}{Purity}&& \multicolumn{4}{c}{F-Score}\\
\cline{4-7}\cline{9-12}\cline{14-17}\cline{19-22} \multicolumn{2}{c}{Feature} && LBP & GIST & HOG & \textbf{MulView} & &  LBP & GIST & HOG & \textbf{MulView} & & LBP & GIST & HOG & \textbf{MulView} & & LBP & GIST & HOG & \textbf{MulView}\\
\hline
\multirow{6}{*}{\begin{sideways}{Single-view Alg.}\end{sideways}}
& $k$-means   &&0.2265 & 0.3085 & 0.2492 & \mycolor{0.3073} && 0.1120 & \textbf{0.1853} & 0.1134 & \mycolor{0.1803} && \textbf{0.2361} & 0.3098 & 0.2439 & \mycolor{0.3133} && 0.1628 & 0.1970 & 0.1363 & \mycolor{0.1996}\\
& $k$-Medoids &&0.1925 & 0.2634 & 0.2268 & \mycolor{0.2605} && 0.0755 & 0.1721 & 0.1298 & \mycolor{0.1461} && 0.1988 & 0.2852 & 0.2329 & \mycolor{0.2690} && 0.1110 & \textbf{0.1973} & 0.1836 & \mycolor{0.1874} \\
& Ak-kmeans    &&0.2159 & 0.2988 & 0.2515 & \mycolor{0.3113} && 0.1000 & 0.1541 & 0.1279 & \mycolor{0.1966} && 0.2255 & 0.2805 & 0.2761 & \mycolor{0.3254} && 0.1662 & 0.1827 & 0.1870 & \mycolor{0.2122} \\
& Nystr\"{o}m &&0.2234 & 0.2459 & 0.2544 & \mycolor{0.2950} && 0.0936 & 0.1222 & 0.1317 & \mycolor{0.1719} && 0.2181 & 0.2585 & 0.2741 & \mycolor{0.3320} && 0.1490 & 0.1749 & 0.1639 & \mycolor{0.2050} \\
& NMF         &&0.2178 & 0.2540 & 0.2509 & \mycolor{0.2737} && 0.1076 & 0.1353 & 0.1434 & \mycolor{0.1610} && 0.2178 & 0.2614 & 0.2705 & \mycolor{0.2887} && 0.1571 & 0.1798 & 0.1609 & \mycolor{0.1854} \\
& LSC-K       &&\textbf{0.2585} & \textbf{0.3192} & 0.2529 & \mycolor{0.3284} && 0.1356 & 0.1806 & 0.1254 & \mycolor{0.2215} && 0.2260 & 0.2660 & 0.2797 & \mycolor{0.3447} && \textbf{0.1748} & 0.1748 & 0.1625 & \mycolor{0.2301} \\
\hline\hline
\multirow{6}{*}{\begin{sideways}{Multi-view Alg.}\end{sideways}}
& AMGL      &&0.2093 & 0.2843 &0.2516  & \mycolor{0.2822} && 0.1131 & 0.1301 & 0.1368 & \mycolor{0.2110} && 0.2149 & 0.3090 & 0.2796 & \mycolor{0.2902} && 0.1311 & 0.1571 & \textbf{0.2021} & \mycolor{0.2305} \\
& MVKM      &&0.2321 & 0.2882 &0.2535  & \mycolor{0.3058} && 0.1181 & 0.1612 & 0.1372 & \mycolor{0.1881} && 0.2115 & 0.3091 & 0.2538 & \mycolor{0.3082} && 0.1461 & 0.1730 & 0.1861 & \mycolor{0.2161} \\
& MLAN      &&0.2109 & 0.2197 &0.2127  & \mycolor{0.3182} && 0.1173 & 0.1255 & 0.1152 & \mycolor{0.1648} && 0.2117 & 0.2258 & 0.2168 & \mycolor{0.3248} && 0.1403 & 0.1614 & 0.1813 & \mycolor{0.1813} \\
& MultiNMF  &&0.2113 & 0.2639 &0.2574  & \mycolor{0.2632} && 0.0986 & 0.1732 & \textbf{0.1605} & \mycolor{0.1708} && 0.2202 & 0.2735 & \textbf{0.2855} & \mycolor{0.2905} && 0.1531 & 0.1789 & 0.1802 & \mycolor{0.1906} \\
& OMVC      &&0.2062 & 0.2706 &0.2544  & \mycolor{0.2739} && 0.1196 & 0.1613 & 0.1222 & \mycolor{0.1744} && 0.1925 & 0.2611 & 0.2592 & \mycolor{0.2637} && 0.1333 & 0.1739 & 0.1761 & \mycolor{0.1885} \\
& MVSC      &&0.2248 & 0.2629 &\textbf{0.2732}  & \mycolor{0.3191} && 0.1293 & 0.1593 & 0.1294 & \mycolor{0.2097} && 0.2132 & 0.3126 & 0.2828 & \mycolor{0.3393} && 0.1481 & 0.1909 & 0.1911 & \mycolor{0.2180}\\
\hline\hline
\multirow{6}{*}{\begin{sideways}{Binary Alg.}\end{sideways}}
& ITQ+$bk$-means \cite{gong2015web} &&0.1861 & 0.2923 &0.2562  & \mycolor{0.3101} && 0.0604 & 0.1746 & 0.1200 & \mycolor{0.2304} && 0.1879 & 0.2842 & 0.2644 & \mycolor{0.3168} && 0.1214 & 0.1954 & 0.1643 & \mycolor{0.2032}\\
& CKM \cite{shen2015cvpr}          &&0.1712 & 0.2382 &0.1906  & \mycolor{0.2794} && 0.0394 & 0.1352 & 0.0738 & \mycolor{0.1823} && 0.1784 & 0.2556 & 0.1962 & \mycolor{0.2844} && 0.1107 & 0.1687 & 0.1389 & \mycolor{0.1990}  \\
& HSIC-TS        &&0.1829 & 0.3030 &0.2523  & \mycolor{0.3568} && 0.1367 & 0.1672 & 0.1013 & \mycolor{0.2376} && 0.1935 & 0.3247 & 0.2577 & \mycolor{0.3665} && 0.1194 & 0.1945 & 0.1525 & \mycolor{0.2309}  \\
& HSIC-F         &&0.1951 & 0.2923 &0.2516  & \mycolor{0.3749} && 0.1289 & 0.1592 & 0.1015 & \mycolor{0.2411} && 0.2062 & 0.3165 &0.2625  & \mycolor{0.3795} && 0.1252 & 0.1832 & 0.1566 & \mycolor{0.2321}\\
& HSIC(ours)     &&0.2275 &0.3128 &0.2597   & \mycolor{\color{blue}{\textbf{0.3865}}} &&\textbf{0.1396} &0.1692 &0.1219 &\mycolor{\color{blue}{\textbf{0.2515}}} &&0.2131 &\textbf{0.3253} &0.2723 &\mycolor{\color{blue}{\textbf{0.3905}}} &&0.1353 &0.1929  & 0.1739 &\mycolor{\color{blue}{\textbf{0.2530}}}\\
\bottomrule
\end{tabular}}\\
\tiny For all single-view methods, features from all views are simply concatenated to obtain the `MulView' results.
\end{center}
\end{table*}

\begin{table*}[!t]
\begin{center}
\caption{\small Time costs (in seconds) of different methods on ImageNet-10.} \label{Table_2}
\resizebox{0.99\textwidth}{!}{
\begin{tabular}{c|cc|cc|cc|cc|cc|cc|cc|cc|cc|cc}
\toprule
\multirow{2}{*}{Alg.}&
\multicolumn{2}{c|}{$k$-means}&\multicolumn{2}{c|}{Ak-kmeans}&\multicolumn{2}{c|}{Nystr\"{o}m}&\multicolumn{2}{c|}{LSC-K}&\multicolumn{2}{c|}{AMGL}&\multicolumn{2}{c|}{MLAN}&\multicolumn{2}{c|}{OMVC}&
\multicolumn{2}{c|}{CKM}&\multicolumn{2}{c|}{HSIC-TS}&
\multicolumn{2}{c}{\textbf{HSIC (ours)}}\\
  & Time & Speedup & Tim. & Speedup & Tim. & Speedup & Tim. & Speedup & Tim. & Speedup & Tim. & Speedup & Tim. & Speedup & Tim. & Speedup & Tim. & Speedup & Tim. & Speedup\\%
\hline
LBP    & 69 & 1$\times$ & 16 & 4.31$\times$ & 15 & 4.60$\times$ & 211 & 0.33$\times$ & 1693 & 0.04$\times$ & 1431 & 0.05$\times$  & 696 & 0.10$\times$ & 17 & 4.06$\times$  &  18 & 3.83 $\times$ & \textbf{4} & \textbf{17.25 $\times$}\\
GIST   & 43 & 1$\times$ & 11 & 3.91$\times$ & 11 & 3.91$\times$ & 226 & 0.19$\times$ & 1730 & 0.03$\times$ & 1557 & 0.03$\times$  & 616 & 0.07$\times$ & 11 & 3.91$\times$  &  16 & 2.69 $\times$ & \textbf{4} & \textbf{10.75 $\times$}\\
HOG    & 82 & 1$\times$ & 11 & 7.46$\times$ & 12 & 6.83$\times$ & 331 & 0.25$\times$ & 1862 & 0.04$\times$ & 2226  & 0.04$\times$ & 643  & 0.13$\times$ & 18 & 4.56$\times$ &  16 & 5.13 $\times$ & \textbf{3} & \textbf{27.33 $\times$}\\
\textbf{MulView} & 201 & 1$\times$& 21 & 9.57$\times$& 19 & 10.58$\times$  & 503 & 0.40$\times$& 3820 & 0.05$\times$ & 3336 & 0.06$\times$ & 1109 & 0.18$\times$ & 27 & 7.44$\times$ & 20 & 10.05 $\times$& \textbf{5} & \color{blue}{\textbf{40.20}} $\times$\\
\bottomrule
\end{tabular}}
\end{center}
\end{table*}

Table \ref{Table_1} demonstrates the performance of all clustering methods. From Table \ref{Table_1}, we can observe in most cases that our HSIC can achieve comparable SVIC results but superior MVIC results in comparison with all the real-valued and binary clustering methods. This indicates the effectiveness of HSIC on the common representation learning and robust cluster structures learning, especially for the MVIC cases. Furthermore, it is clear that HSIC is superior to HSIC-F and HSIC-ST, which demonstrates the robustness and effectiveness of the joint learning framework.

The computational costs are illustrated in Table \ref{Table_2}. From its last three columns, we can see that the binary clustering methods can reduce the computational time compared with the real-valued ones such as $k$-means and LSC-K, due to the highly efficient distance calculation in the Hamming space. Particularly, our HSIC is much faster than the compared real-valued and binary clustering methods, which also proves the superiority of the developed efficient optimization algorithm. Specifically, the speed-up of our HSIC for MVIC is very clear by a margin of $40.20$ times compared to $k$-means. For memory footprint, $k$-means and our HSIC respectively require $361$ MB and $2.73$ MB, \textit{i.e.}, $ \approx 132$ times memory can be reduced using HSIC.

\begin{table*}[!t]
\begin{center}
\newcommand{\tabincell}[2]{\begin{tabular}{@{}#1@{}}#2\end{tabular}}
\caption{\small Performance comparisons on the three large-scale datasets. The bold \textbf{black} and {\color{blue}{\textbf{blue}}} numbers indicate the best single-view and multi-view results, respectively.}
\label{Table_3}
\resizebox{0.99\textwidth}{!}{
\begin{tabular}{c|c|c|ccccccccccc}
\hline
&Metric&Alg. & $k$-means & $k$-mean++ & $k$-Medoids & Ak-kmeans & LSC-K & Nystr\"{o}m & \tabincell{c}{ITQ+\\$bk$-means} & CKM & HSIC-TS & HSIC-F & \textbf{HSIC}\\ %
\hline
\multirow{16}{*}{\begin{sideways}{Cifar-10}\end{sideways}}
&\multirow{4}{*}{ACC}
&LBP       & 0.2185 & 0.2182  & 0.2171 & 0.2066 & 0.2550 & 0.2339 & 0.2322 &0.2225 &0.2440 &0.2536 &\textbf{0.2681}\\
&&GIST     & 0.2842 & 0.2845  & 0.2419 & 0.2847 & 0.3010 & 0.2592 & 0.2777 &0.2521 &0.3209 &0.3456 &\textbf{0.3595}\\
&&HOG      & 0.2661 & 0.2703  & 0.2456 & 0.2608 & 0.2838 & 0.2408 & 0.2481 &0.2294 &0.3178 &0.3394 &\textbf{0.3389}\\
&&\textbf{MulView} & \mycolor{0.2877} & \mycolor{0.2882} & \mycolor{0.2630} & \mycolor{0.2879} & \mycolor{0.3488} & \mycolor{0.2747} & \mycolor{0.2787} &\mycolor{0.2703} &\mycolor{0.3742} &\mycolor{0.3809} &\mycolor{\color{blue}{\textbf{0.3951}}}\\
\cline{2-14}
&\multirow{4}{*}{NMI}
&LBP       & 0.1044 & 0.1044 & 0.0862 & 0.1021 & 0.1303 & 0.0922 & 0.0963 &0.1092 &0.1105 &0.1094 &\textbf{0.1220}\\
&&GIST     & 0.1692 & 0.1691 & 0.1238 & 0.1692 & 0.1869 & 0.1226 & 0.1502 &0.1184 &0.2063 &0.2134 &\textbf{0.2299}\\
&&HOG      & 0.1634 & 0.1645 & 0.1328 & 0.1607 & 0.1668 & 0.1415 & 0.1570 &0.1034 &0.2053 &\textbf{0.2199} &0.2170\\
&&\textbf{MulView} & \mycolor{0.1803} & \mycolor{0.1805} & \mycolor{0.1565} & \mycolor{0.1808} & \mycolor{0.2382} & \mycolor{0.1511} & \mycolor{0.1613} &\mycolor{0.1499} &\mycolor{0.2547} &\mycolor{0.2596} &\mycolor{\color{blue}{\textbf{0.2629}}}\\
\cline{2-14}
&\multirow{4}{*}{Purity}
&LBP       & 0.2401 & 0.2400 & 0.2339 & 0.2275 & 0.2768 & 0.2445 & 0.2490 &0.2476 &0.2526 &0.2697 &\textbf{0.2837}\\
&&GIST     & 0.3056 & 0.3052  & 0.2483 & 0.3054 & 0.3306 & 0.2626 & 0.2882 &0.2649 &0.3650 &0.3651 &\textbf{0.3828}\\
&&HOG      & 0.2943 & 0.2953 & 0.2561 & 0.2847 & 0.3039 & 0.2655 & 0.2756 &0.2319 &0.3199 &\textbf{0.3589} &0.3481\\
&&\textbf{MulView} & \mycolor{0.3136} & \mycolor{0.3138} & \mycolor{0.2921} & \mycolor{0.3148} & \mycolor{0.3787} & \mycolor{0.2975} & \mycolor{0.2953} &\mycolor{0.2846} &\mycolor{0.3956} &\mycolor{0.4045} &\mycolor{\color{blue}{\textbf{0.4204}}}\\
\cline{2-14}
&\multirow{4}{*}{F-score}
&LBP       & 0.1677 & 0.1676 & 0.1703 & 0.1643 & 0.1692 & 0.1517 & 0.1685 &0.1509 &0.1717 &0.1670 &\textbf{0.1721}\\
&&GIST     & 0.1866 & 0.1866  & 0.1744 & 0.1867 & 0.2044 & 0.1654 & 0.1808 &0.1606 &0.2318 &0.2318 &\textbf{0.2397}\\
&&HOG      & 0.1887 & 0.1895 & 0.1808 & 0.1882 & 0.1878 & 0.1680 & 0.1769 &0.1479 &0.2221 &0.2337 &\textbf{0.2383}\\
&&\textbf{MulView} & \mycolor{0.1998} & \mycolor{0.2001} & \mycolor{0.2035} & \mycolor{0.2001} & \mycolor{0.2477} & \mycolor{0.1793} & \mycolor{0.1863} &\mycolor{0.1807} &\mycolor{0.2422} &\mycolor{0.2564} &\mycolor{\color{blue}{\textbf{0.2595}}}\\
\cline{2-14}
\hline \hline
\multirow{16}{*}{\begin{sideways}{YouTube-Faces (YTBF)}\end{sideways}}
&\multirow{4}{*}{ACC}
&LBP       & 0.5870 & 0.5994  & 0.5262 & 0.5584 & 0.6017 & 0.5647 & 0.5765 &0.5319 &0.5930 &0.6208 &\textbf{0.6471}\\
&&GIST     & 0.4081 & 0.4068  & 0.3584 & 0.2937 & 0.4638 & 0.4497 & 0.3547 &0.3760 &0.5432 &0.6059 &\textbf{0.6121}\\
&&HOG      & 0.5751 & 0.5821  & 0.4810 & 0.5542 & 0.5830 & 0.5642 & 0.5574 &0.5584 &0.5436 &\textbf{0.6133} &0.6099\\
&&\textbf{MulView} & \mycolor{0.5927} & \mycolor{0.6067}  & \mycolor{0.5290} & \mycolor{0.5562} & \mycolor{0.6099} & \mycolor{0.6190} & \mycolor{0.5852} &\mycolor{0.5574} &\mycolor{0.5974} &\mycolor{0.6315} &\mycolor{\color{blue}{\textbf{0.6547}}}\\
\cline{2-14}
&\multirow{4}{*}{NMI}
&LBP       & 0.7473 & 0.7460  & 0.6835 & 0.7251 &\textbf{0.7725} & 0.7515 & 0.6870 &0.6222 &0.7256 &0.7478 &0.7690\\
&&GIST     & 0.5528 & 0.5472  & 0.5062 & 0.4165 & 0.6237 & 0.6630 & 0.5146 &0.5094 &0.6889 &0.7272 &\textbf{0.7436}\\
&&HOG      & 0.7442 & 0.7375  & 0.6640 & 0.7206 & \textbf{0.7536} & 0.7193 & 0.6827 &0.6805 &0.6965 &0.7342 &0.7483\\
&&\textbf{MulView} & \mycolor{0.7492} & \mycolor{0.7488}  & \mycolor{0.6774} & \mycolor{0.7215} & \mycolor{0.7515} & \mycolor{0.7307} & \mycolor{0.6921} &\mycolor{0.6827} &\mycolor{0.7579} &\mycolor{0.7785} &\mycolor{\color{blue}{\textbf{0.7899}}}\\
\cline{2-14}
&\multirow{4}{*}{Purity}
&LBP      & 0.6744 & 0.6760  & 0.6033 & 0.6155 & 0.6782 & 0.6697 & 0.6529 &0.5695 &0.6597 &0.6600 &0.6915\\
&&GIST    & 0.4641 & 0.4622  & 0.4315 & 0.3157 & 0.5366 & 0.5729 & 0.4405 &0.4398 &0.6099 &0.6530 &0.6766\\
&&HOG     & 0.6499 & 0.6481  & 0.5733 & 0.6218 & 0.6602 & 0.6602 & 0.6257 &0.6369 &0.6105 &0.6606 &0.6682\\
&&\textbf{MulView} & \mycolor{0.6712} & \mycolor{0.6692}  & \mycolor{0.5969} & \mycolor{0.6376} & \mycolor{0.6687} & \mycolor{0.6778} & \mycolor{0.6642} &\mycolor{0.6257} &\mycolor{0.6615} &\mycolor{0.6955} &\mycolor{\color{blue}{\textbf{0.7023}}}\\
\cline{2-14}
&\multirow{4}{*}{F-score}
&LBP       & 0.4240 & 0.4378  & 0.4034 & 0.4412 & 0.5058 & 0.4375 & 0.4421 &0.4105 &0.4286 &0.4982 &\textbf{0.5123}\\
&&GIST     & 0.2567 & 0.2551  & 0.2310 & 0.1666 & 0.3390 & 0.3455 & 0.2308 &0.2578 &0.3367 &0.4871 &\textbf{0.4914}\\
&&HOG      & 0.4813 & 0.4572  & 0.3715 & 0.4464 & 0.4627 & 0.3990 & 0.4303 &0.4663 &0.3379 &0.4960 &\textbf{0.5016}\\
&&\textbf{MulView} & \mycolor{0.4886} & \mycolor{0.4853}  & \mycolor{0.4236} & \mycolor{0.4209} & \mycolor{0.4650} & \mycolor{0.4211} & \mycolor{0.4650} &\mycolor{0.4303} &\mycolor{0.4517} &\mycolor{0.5113} &\mycolor{\color{blue}{\textbf{0.5425}}}\\
\cline{2-14}
\hline \hline
\multirow{24}{*}{\begin{sideways}{NUS-WIDE}\end{sideways}}
&\multirow{4}{*}{ACC}
&CH      & 0.1321 & 0.1370  & \textbf{0.1433} & 0.1351 & 0.1253 & 0.1391 & 0.1193 &0.1244 &0.1243 &0.1314 &0.1282\\
&&CM     & 0.1334 & \textbf{0.1379}  & 0.1305 & 0.1300 & 0.1297 & 0.1130 & 0.1123 &0.1202 &0.1346 &0.1376 &0.1360\\
&&CORR   & 0.1352 & \textbf{0.1358}  & 0.1222 & 0.1301 & 0.1344 & 0.1277 & 0.1143 &0.1161 &0.1349 &0.1253 &0.1279\\
&&EDH    & 0.1402 & \textbf{0.1425}  & 0.1382 & 0.1399 & 0.1266 & 0.1129 & 0.1180 &0.1223 &0.1343 &0.1343 &0.1396\\
&&WT     & 0.1145 & 0.1182  & 0.1176 & 0.1169 & 0.1110 & 0.1226 & 0.1240 &0.1172 &0.1242 &0.1147 &\textbf{0.1293}\\
&&\textbf{MulView} & \mycolor{0.1434} & \mycolor{0.1458}  & \mycolor{0.1545} & \mycolor{0.1499} & \mycolor{0.1567} & \mycolor{0.1452} & \mycolor{0.1295} &\mycolor{0.1296} &\mycolor{0.1607} &\mycolor{0.1639} &\mycolor{\color{blue}{\textbf{0.1661}}}\\
\cline{2-14}%
&\multirow{4}{*}{NMI}
&CH      & 0.0687 & 0.0675  & 0.0706 & 0.0682 & 0.0638 & 0.0684 & 0.0629 &0.0613 &0.0668 &0.0662 &\textbf{0.0938}\\
&&CM     & 0.0755 & 0.0687  & 0.0615 & 0.0747 & 0.0746 & 0.0656 & 0.0625 &0.0580 &0.0775 &0.0870 &\textbf{0.0944}\\
&&CORR   & 0.0701 & 0.0699  & 0.0639 & 0.0714 & 0.0691 & 0.0661 & 0.0655 &0.0589 &0.0784 &0.0652 &\textbf{0.0882}\\
&&EDH    & 0.0844 & 0.0877  & 0.0830 & 0.0900 & 0.0866 & 0.0707 & 0.0758 &0.0731 &\textbf{0.0961} &0.0872 &0.0925\\
&&WT     & 0.0571 & 0.0593  & 0.0559 & 0.0558 & 0.0661 & 0.0711 & 0.0632 &0.0645 &0.0878 &0.0652 &\textbf{0.0748}\\
&&\textbf{MulView} & \mycolor{0.0944} & \mycolor{0.0967}  & \mycolor{0.0823} & \mycolor{0.0947} & \mycolor{0.0980} & \mycolor{0.0880} & \mycolor{0.0773} &\mycolor{0.0696} &\mycolor{0.0937} &\mycolor{0.0989}  &\mycolor{\color{blue}{\textbf{0.1032}}}\\
\cline{2-14}
&\multirow{4}{*}{Purity}
&CH      & 0.2459 & 0.2418  & 0.2498 & 0.2439 & 0.2432 & 0.2443 & 0.2422 &0.2390 &0.2437 &0.2397 &\textbf{0.2589}\\
&&CM     & 0.2453 & 0.2459  & 0.2284 & 0.2507 & 0.2516 & 0.2495 & 0.2433 &0.2414 &\textbf{0.2601} &0.2371 &0.2515\\
&&CORR   & 0.2370 & 0.2341  & 0.2402 & 0.2413 & 0.2408 & 0.2387 & 0.2404 &0.2344 &0.2564 &0.2337 &\textbf{0.2589}\\
&&EDH    & 0.2388 & 0.2448  & 0.2365 & 0.2467 & 0.2393 & 0.2193 & 0.2354 &0.2308 &0.2451 &0.2296 &\textbf{0.2587}\\
&&WT     & 0.2256 & 0.2274  & 0.2235 & 0.2237 & 0.2297 & 0.2328 & 0.2273 &0.2256 &0.2339 &0.2306 &\textbf{0.2393}\\
&&\textbf{MulView} & \mycolor{0.2625} & \mycolor{0.2634} & \mycolor{0.2446} & \mycolor{0.2711} & \mycolor{0.2657} & \mycolor{0.2546} & \mycolor{0.2487} & \mycolor{0.2413} & \mycolor{0.2647} &\mycolor{0.2653} &\mycolor{\color{blue}{\textbf{0.2753}}}\\
\cline{2-14}
&\multirow{4}{*}{F-score}
&CH      & 0.1128 & 0.1134  & \textbf{0.1147} & 0.1095 & 0.0946 & 0.1031 & 0.0867 &0.0882 &0.0863 &0.0901 &0.1009\\
&&CM     & 0.1011 & \textbf{0.1128}  & 0.0981 & 0.0956 & 0.0896 & 0.0867 & 0.0836 &0.0879 &0.0941 &0.1095 &0.1010\\
&&CORR   & 0.1005 & \textbf{0.1027}  & 0.0954 & 0.0947 & 0.0945 & 0.0969 & 0.0854 &0.0841 &0.0985 &0.0888 &0.0965\\
&&EDH    & \textbf{0.1163} & 0.1150  & 0.1079 & 0.1149 & 0.0972 & 0.0865 & 0.0892 &0.0899 &0.0966 &0.1130 &0.1033\\
&&WT     & 0.0933 & 0.0949  & 0.0940 & 0.0975 & 0.0893 & 0.0914 & 0.0889 &0.0892 &0.0903 &0.0912 &\textbf{0.1019}\\
&&\textbf{MulView} & \mycolor{0.1106} & \mycolor{0.1125}  & \mycolor{0.1105} & \mycolor{0.1061} & \mycolor{0.1071} & \mycolor{0.1006} & \mycolor{0.0905} &\mycolor{0.0903} &\mycolor{0.1076} &\mycolor{0.1055} &\mycolor{\color{blue}{\textbf{0.1216}}}\\
\hline
\end{tabular}}
\\
\tiny For all single-view methods, features from all views are simply concatenated to obtain the `MulView' results.
\end{center}
\end{table*}

\begin{figure}[!b]
\centering\resizebox{.92\textwidth}{.24\textheight}{
\includegraphics[]{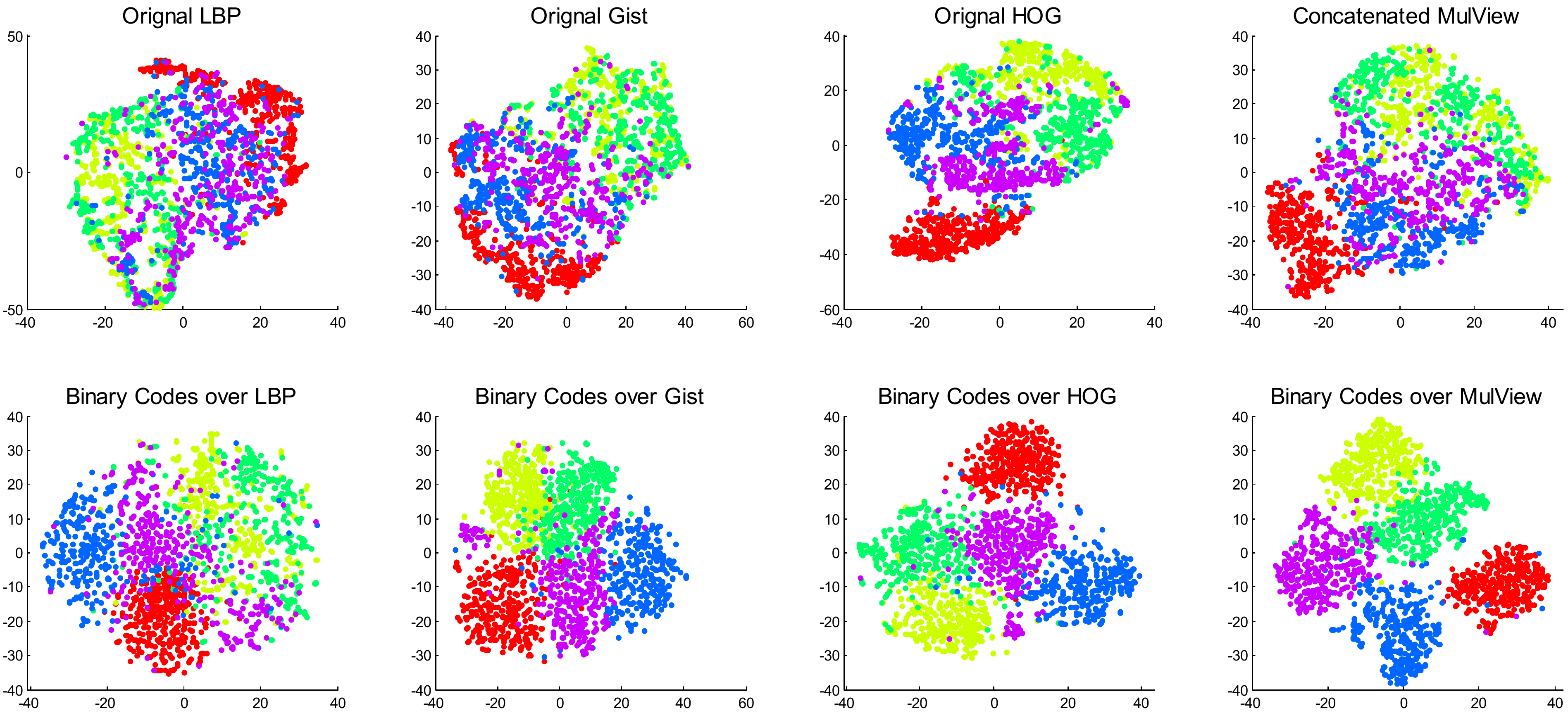}}
\caption{\small $t$-SNE visualization of randomly selected 5 classes from ImageNet-10. The two rows show the real-valued features and 128-bit HSIC-based binary codes, respectively.}\label{fig_2}
\end{figure}

\textbf{Why does HSIC Outperform the Real-Valued Methods?} Table \ref{Table_1} clearly shows that HSIC achieves competitive or superior clustering performance compared to the real-valued clustering methods. The favorable performance mainly comes from: \textbf{1)} HSIC greatly benefits from the proposed effective discrete optimization algorithm such that the learned binary representations can eliminate some redundant and noisy information in the original real-valued features. As can be seen in Fig. \ref{fig_2}, the similarity structures of the same clusters are enhanced in the coding space, meanwhile, some disturbances from the original features are excluded to refine the learned representation. \textbf{2)} For image clustering, binary features are more robust to local changes since small variations caused by varying environments can be eliminated by quantized binary codes. \textbf{3)} HSIC is a unified interactive learning framework of the optimal binary codes and clustering structures, which is shown to be better than those disjoint learning approaches (\textit{e.g.}, LSC-K, NMF, MVSC, AMGL and MLAN).

\begin{table*}[!t]
\begin{center}
\caption{Time costs (in seconds) on the three large-scale multi-view datasets.} \label{Table_4}
\resizebox{0.95\textwidth}{!}{
\begin{tabular}{c|c|cc|cc|cc|cc|cc|cc|cc|cc|cc}
\hline
&\multirow{2}{*}{Alg.}&
\multicolumn{2}{c|}{$k$-means}&
\multicolumn{2}{c|}{$k$-means++}&
\multicolumn{2}{c|}{Ak-kmeans}&
\multicolumn{2}{c|}{LSC-K}&
\multicolumn{2}{c|}{Nystr\"{o}m}&
\multicolumn{2}{c|}{ITQ+$bk$-means}&
\multicolumn{2}{c|}{CKM}&
\multicolumn{2}{c|}{HSIC-TS}&
\multicolumn{2}{c}{\textbf{HSIC (ours)}}\\
\cline{3-20}
  && Time & Speedup & Time & Speedup & Time & Speedup & Time & Speedup & Time & Speedup & Time & Speedup & Time & Speedup & Time & Speedup & Time & Speedup\\
\hline
\multirow{4}{*}{\begin{sideways}{Cifar-10}\end{sideways}}
&LBP     & 409 & 1$\times$ & 294 & 1.39$\times$ & 61 &  6.71$\times$    & 112 & 3.65$\times$ & 26 &  15.73$\times$ & 24 & 17.04$\times$ & 29 & 14.10$\times$ & 29 &  14.10$\times$ & \mycolor{\textbf{10}} &  \mycolor{\textbf{40.90$\times$}}\\
&GIST    & 305 & 1$\times$ & 334 & 0.91$\times$ & 56 &  5.44$\times$    & 834 & 0.37$\times$ & 28 &   10.89$\times$ & 23 & 13.26$\times$ & 28 & 10.89$\times$ & 30 &  10.17$\times$ & \mycolor{\textbf{10}} &  \mycolor{\textbf{30.50$\times$}}\\
&HOG     & 412 & 1$\times$ & 266 & 1.55$\times$ & 58 &  7.10$\times$    & 913 & 0.45$\times$ & 32 &   12.87$\times$ & 27 & 15.26$\times$ & 30 & 13.73$\times$ & 25 &  16.48$\times$ & \mycolor{\textbf{10}} &  \mycolor{\textbf{41.20$\times$}}\\
&MulView & 977 & 1$\times$ & 791 & 1.23$\times$ & 77 &  12.69$\times$   & 1877 & 0.52$\times$ & 58 &  16.85$\times$ & 48 & 20.35$\times$ & 46 & 21.24$\times$ & 34 &  28.74$\times$ & \mycolor{\textbf{17}} & \mycolor{\textbf{57.47$\times$}}\\
\hline\hline
\multirow{4}{*}{\begin{sideways}{YTBF}\end{sideways}}
&LBP     & 2344 & 1$\times$ & 1974 & 1.18$\times$ & 533 &  4.40$\times$  & 3546 & 0.66$\times$ & 766 &   3.06$\times$ & 90  & 26.04$\times$ & 141 & 16.62$\times$ & 97 & 24.17$\times$ & \mycolor{\textbf{40}} & \mycolor{\textbf{58.60$\times$}}\\
&GIST    & 2299 & 1$\times$ & 1705 & 1.34$\times$ & 515 &  4.46$\times$  & 3796 & 0.61$\times$ & 828 &   2.78$\times$ & 107 & 21.49$\times$ & 153 & 15.03$\times$ & 98 &  23.46$\times$ & \mycolor{\textbf{36}} & \mycolor{\textbf{63.86$\times$}}\\
&HOG     & 3329 & 1$\times$ & 1508 & 2.21$\times$ & 523 &  6.37$\times$  & 4042 & 0.83$\times$ & 870 &   3.83$\times$ & 104 & 32.01$\times$ & 197 & 16.90$\times$ & 105 &  31.71$\times$ & \mycolor{\textbf{48}} &  \mycolor{\textbf{69.35$\times$}}\\
&MulView & 5879 & 1$\times$ & 4250 & 1.38$\times$ & 539 & 10.91$\times$  &12546 & 0.47$\times$ & 998 &   5.89$\times$ & \textbf{110} & \textbf{53.45}$\times$ & 309 & 19.03$\times$ & 162 & 36.29$\times$ & \mycolor{139} &  \mycolor{42.30$\times$}\\
\hline\hline
\multirow{6}{*}{\begin{sideways}{NUS-WIDE}\end{sideways}}
&CH     & 1027 & 1$\times$ & 852 & 1.21$\times$ & 464 &  2.21$\times$    & 1693 & 0.61$\times$ & 327 &   3.14$\times$ & 91 & 11.29$\times$ & 83 & 12.37$\times$ & 85 &  12.08$\times$ & \mycolor{\textbf{34}} &  \mycolor{\textbf{30.21$\times$}}\\
&CM     & 1206 & 1$\times$ & 937 & 1.29$\times$ & 464 &  2.60$\times$    & 1987 & 0.61$\times$ & 352 &   3.43$\times$ & 82 & 14.71$\times$ & 93 & 12.97$\times$ & 89 &  13.55$\times$ & \mycolor{\textbf{35}} &  \mycolor{\textbf{34.46$\times$}}\\
&CORR   & 1101 & 1$\times$ & 876 & 1.26$\times$ & 467 &  2.36$\times$    & 1854 & 0.59$\times$ & 382 &   2.88$\times$ & 83 & 13.27$\times$ & 83 & 13.26$\times$ & 89 &  12.37$\times$ & \mycolor{\textbf{35}} & \mycolor{\textbf{31.46$\times$}}\\
&EDH    & 1000 & 1$\times$ & 829 & 1.21$\times$ & 454 &  2.21$\times$    & 1825 &  0.55$\times$ & 371 &  2.70$\times$ & 99 & 10.10$\times$ & 91 & 10.99$\times$ & 98 &  10.20$\times$ & \mycolor{\textbf{34}} & \mycolor{\textbf{29.41$\times$}}\\
&WT     & 1206 & 1$\times$ & 784 & 1.54$\times$ & 491 &  2.46$\times$    & 1984 & 0.61$\times$ & 427 &   2.82$\times$ & 82 & 14.71$\times$ & 99 & 12.18$\times$ & 81 &  14.89$\times$ & \mycolor{\textbf{34}} &  \mycolor{\textbf{35.47$\times$}}\\
&MulView& 1711 & 1$\times$ & 1147& 1.49$\times$ & 479 &  3.57$\times$    & 8978 & 0.19$\times$ & 485 &   3.53$\times$ & 105& 16.30$\times$ & 142& 12.05$\times$ & 112&  15.28$\times$ & \mycolor{\textbf{81}} &  \mycolor{\textbf{21.12$\times$}}\\
\hline
\end{tabular}}
\end{center}
\end{table*}

\begin{table}[!t]
\newcommand{\tabincell}[2]{\begin{tabular}{@{}#1@{}}#2\end{tabular}}
\begin{center}
\caption{\small Memory footprint of `MulView' $k$-means and HSIC on the three large-scale datasets. `Reduction' denotes the times of memory reduction against $k$-means.}
\label{Table_5}
\resizebox{0.95\textwidth}{!}{
\begin{tabular}{ccccc|ccccccccc}
\toprule
\multirow{2}{*}{Datasets}&
\multicolumn{3}{c}{Memory w.r.t. $k$-means}&&&
\multicolumn{4}{c}{Memory w.r.t. \textbf{HSIC (ours)}}\\
\cline{2-4}\cline{7-10} & \tabincell{c}{Data\\(Real-valued features)} & Centroids & Reduction &&& \tabincell{c}{Data\\(128-bit binary codes)} & Centroids& Projection & Reduction \\
\hline
Cifar-10 (60,000 images)     &  1.62GB &  0.28MB  & 1$\times$ &&& 0.92MB &  0.15$\times 10^{-3}$MB & 2.53MB &\textbf{481$\times$}\\
YTBF (182,881 images)         &  4.94GB &  2.46MB  & 1$\times$ &&&  2.79MB &  1.36$\times 10^{-3}$MB & 2.53MB & \textbf{951$\times$}\\
NUS-WIDE (195,834 images)     &  961MB &  0.10MB  & 1$\times$ &&&  2.99MB &  0.32$\times 10^{-3}$MB & 2.53MB & \textbf{174$\times$} \\
\bottomrule
\end{tabular}}
\end{center}
\end{table}

\subsection{Experiments on Large-Scale Datasets}\label{ExLarge}
To show the strong scalability of HSIC on the large-scale MVIC problem, we compare HSIC with several state-of-the-art scalable clustering methods on three large-scale multi-view datasets. The clustering performance is summarized in Table \ref{Table_3}. Given these results, we have the following observations: \textbf{1)} Generally, MVIC performs better than SVIC, which implies the necessity of incorporating complementary traits of multiple features for image clustering. Particularly, our HSIC achieves competitive or better SVIC results but consistent best MVIC performance. This mainly owes to the adaptive weights learning strategy and the exploiting of sharable and individual information from heterogeneous features. \textbf{2)} From the last three columns of Table \ref{Table_3}, we can observe that HSIC and its variants tend to be better than the real-valued ones. This shows that the binary codes learned by HISC are competitive to the real-valued ones. \textbf{3)} When comparing to HSIC-TS and HSIC-F, HSIC in most cases achieves superior performance. This further reflects the advantages of the unified learning strategy and robust binary cluster structure construction.

The comparisons of running time and memory footprint are illustrated in Tables \ref{Table_4} and \ref{Table_5}, respectively. From Table \ref{Table_4}, we can observe that our HSIC is the fastest method in most cases. Table \ref{Table_5} shows that HSIC significantly reduces the memory load for large-scale MVIC compared to $k$-means. The memory cost of HSIC is similar to other binary clustering methods but clearly less than the real-valued methods. Moreover, as shown in Tables \ref{Table_4} and \ref{Table_5}, for MVIC on NUS-WIDE with $5$ views, HSIC can cluster near one million ($195,834\times 5$) features in $81$ seconds using only $5.52$ MB memory, while $k$-means needs about $29$ minutes with $961$ MB memory. Thus, HSIC can effectively address large-scale MVIC with much less computational time and memory footprint.

\subsection{Empirical Analysis}
\begin{figure}[!t]
\centering
\subfloat[ACC]
{\includegraphics[width=1.10in]{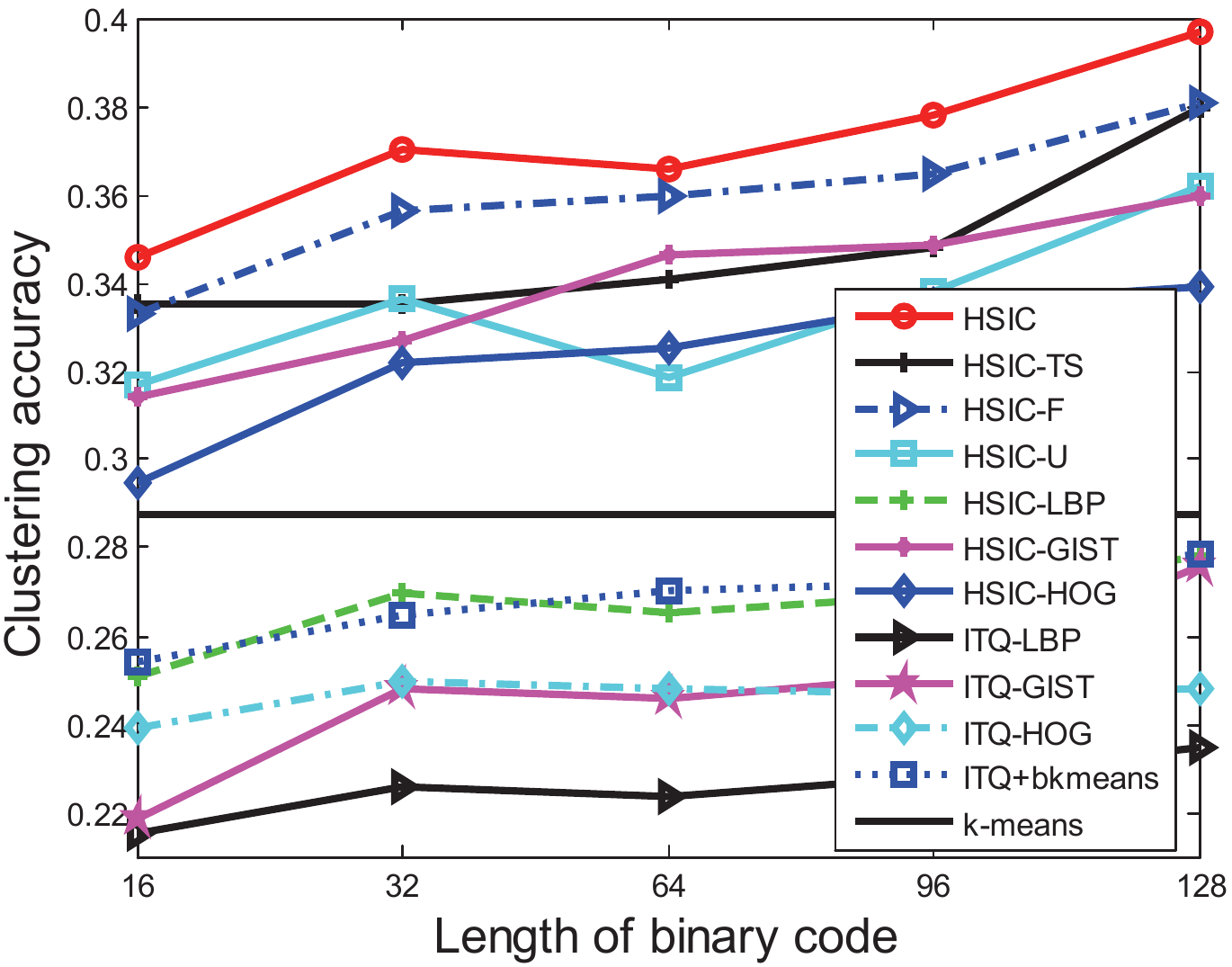}}~
\subfloat[NMI]
{\includegraphics[width=1.10in]{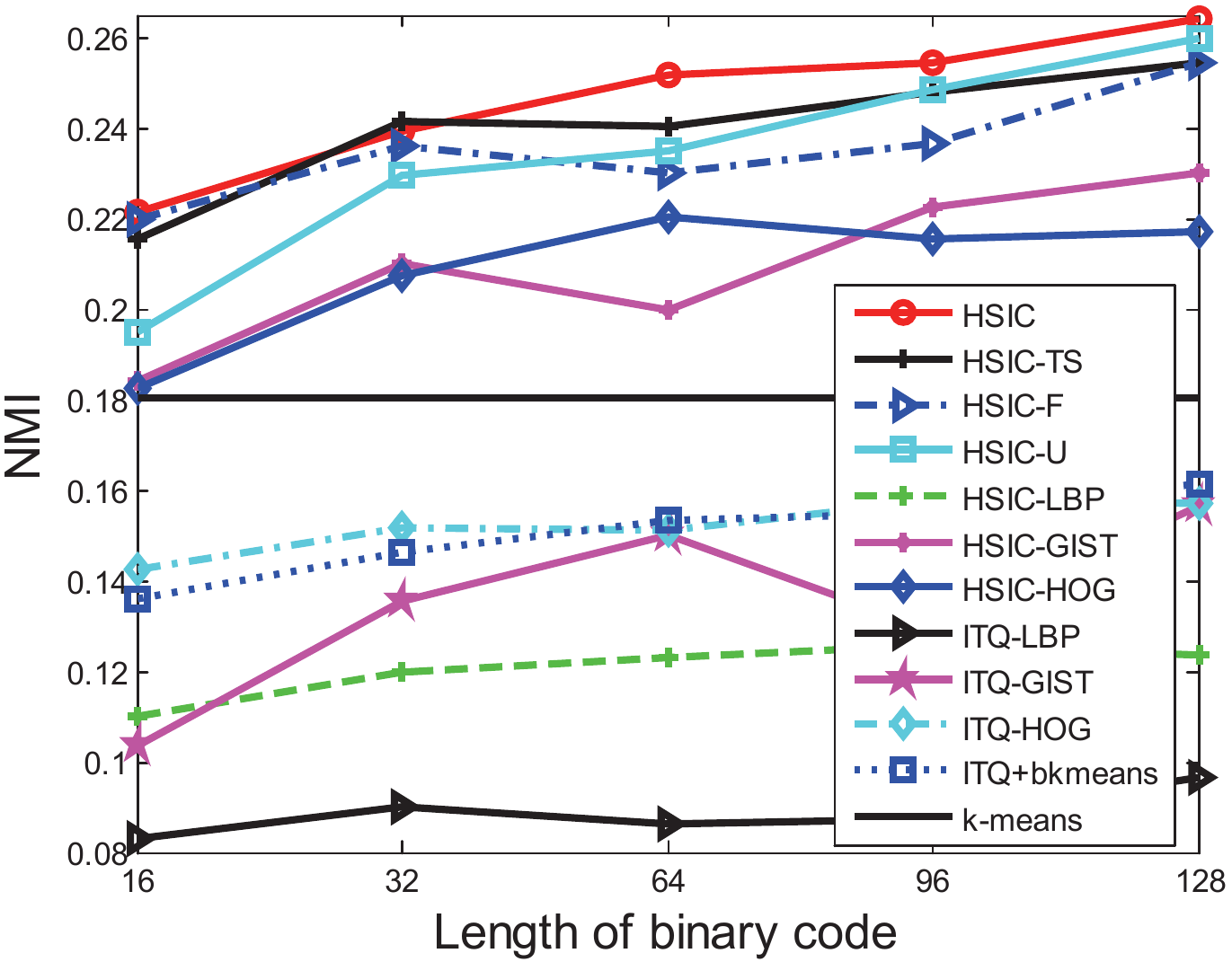}}~
\subfloat[Purity]
{\includegraphics[width=1.10in]{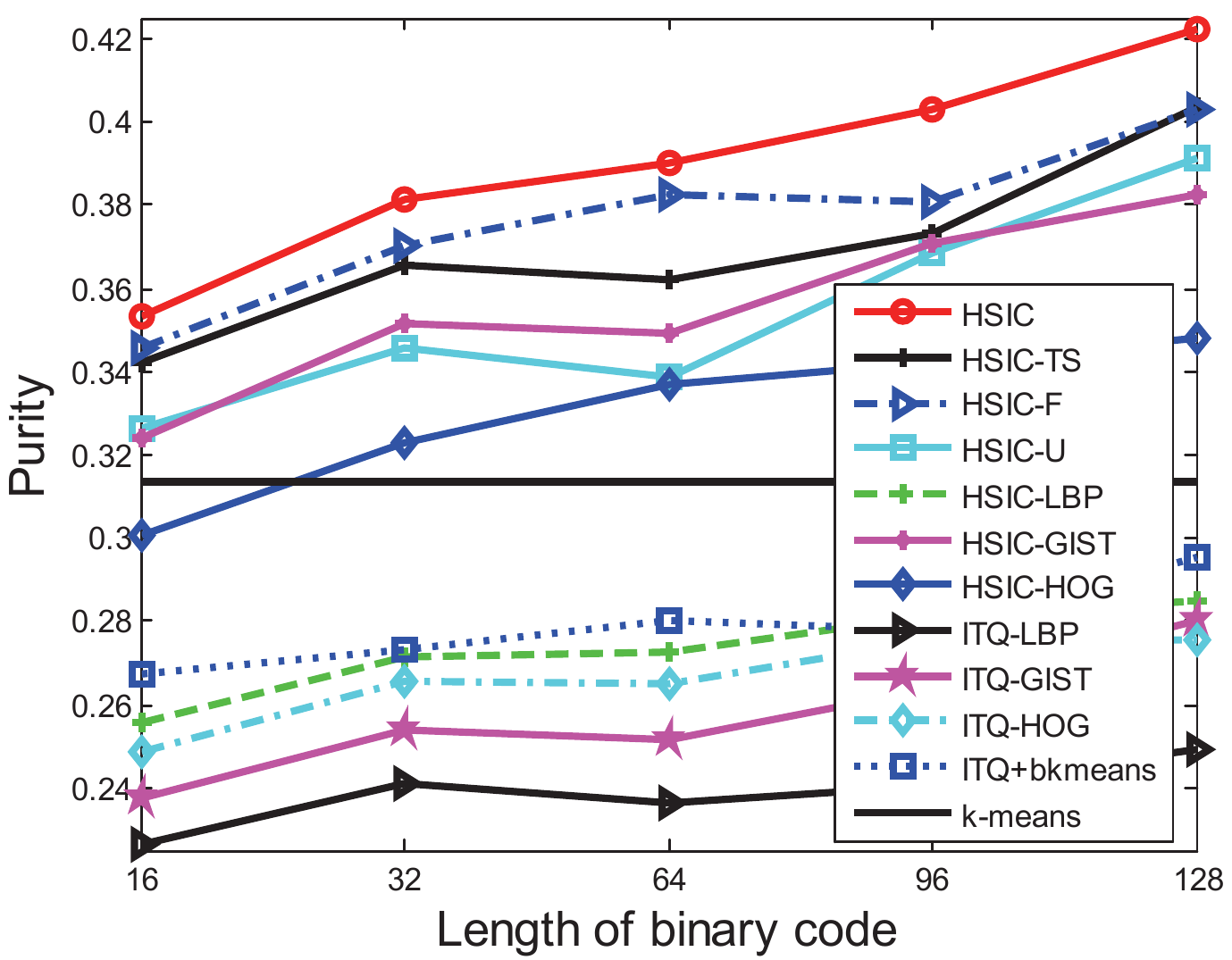}}~
\subfloat[F-Score]
{\includegraphics[width=1.10in]{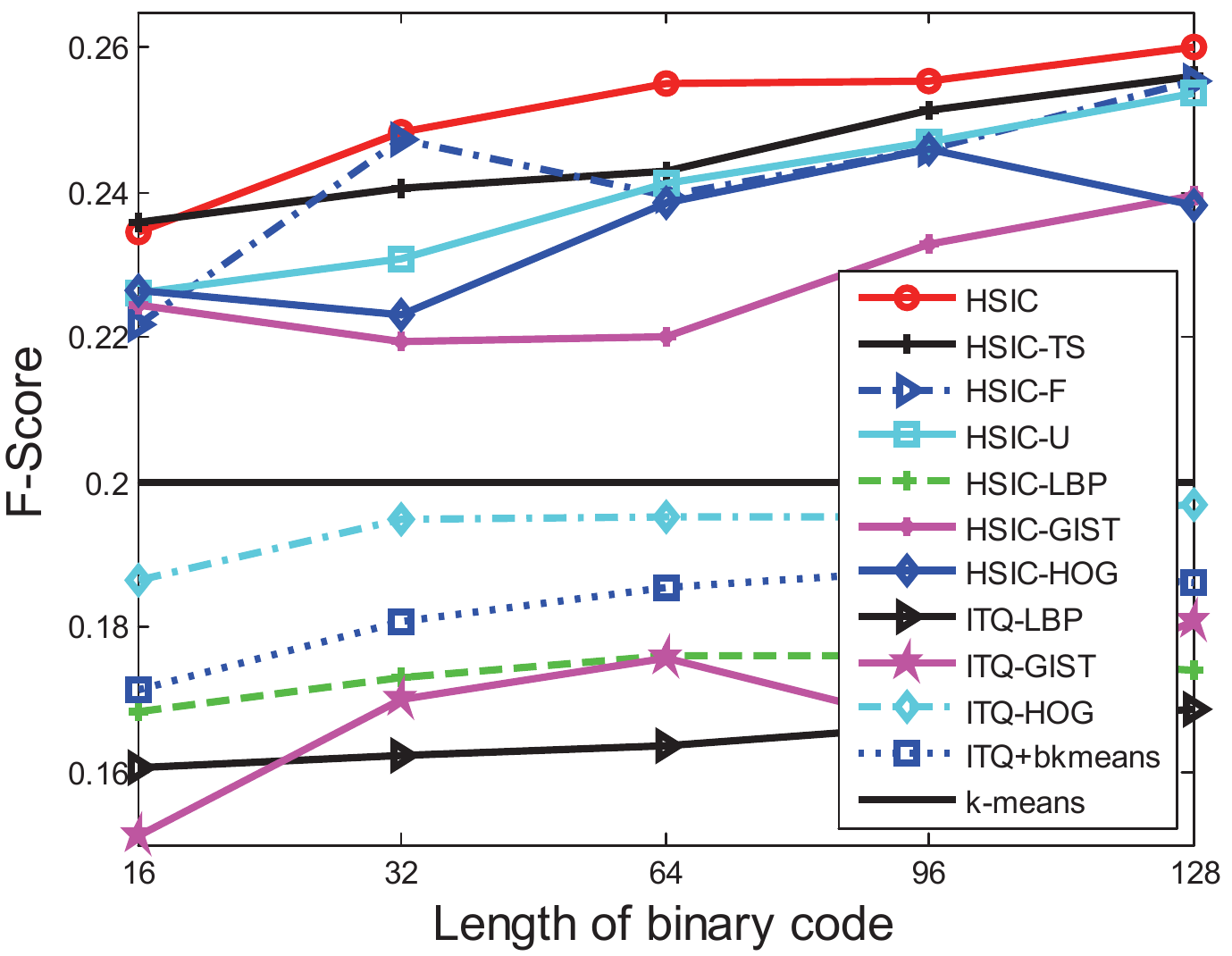}}
\caption{\small Performance of different clustering methods vs. code lengths on Cifar-10.}
\label{fig_3}
\end{figure}

\textbf{Component Analysis}: We evaluate the effectiveness of different components of HSIC in Fig. \ref{fig_3}. Specifically, in addition to `HSIC-TS' and `HSIC-F', we have `HSIC-U' by removing the balanced and independence constraints on binary codes and clustering centroids. HSIC-`view' and ITQ-`view' respectively refer to the SVIC results obtained using HSIC and ITQ+$bk$-means on the `view'-specific features. From Fig. \ref{fig_3}, we can observe that each component contributes essentially to the enhanced performance, and lacking any component will deteriorate the performance.

\textbf{Effect of Code Length}: We show our performance changes with the increasing code lengths in Fig. \ref{fig_3}. In general, longer codes may provide more information for higher clustering performance. Specifically, both ITQ and HSIC based methods tend to achieve improved performance with increasing numbers of bits. Moreover, HSIC-based methods are superior to the baseline $k$-means when the code length is larger than $32$. The best clustering results are established by HSIC w.r.t. different code lengths, because HSIC can effectively coordinate the importance of different views and mine the semantic correlations between them.

\begin{figure}[!t]
\centering
\resizebox{0.99\textwidth}{!}{
\includegraphics[]{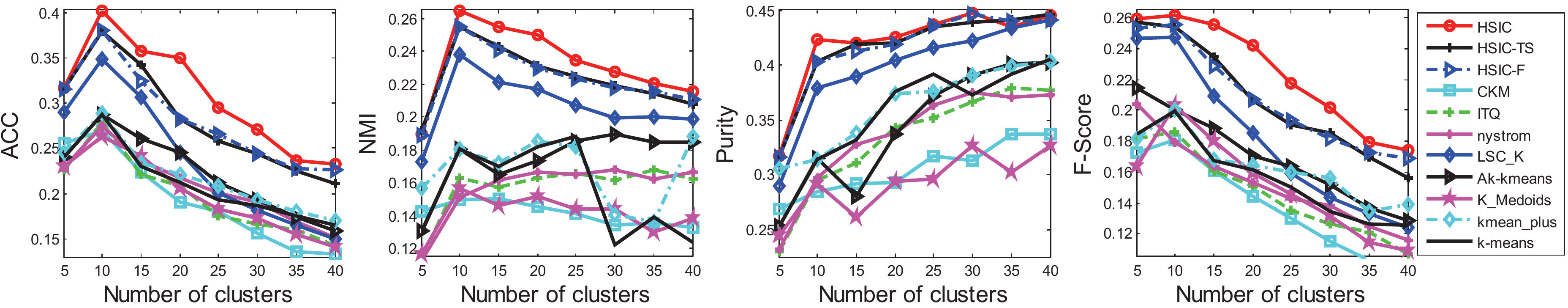}}
\caption{\small Performance of different clustering methods vs. numbers of clusters on Cifar-10.}
\label{fig_4}
\end{figure}

\textbf{Effect of Number of Clusters}: All the above experiments are evaluated based on the ground-truth cluster numbers. However, if the number of clusters is unknown, how will the performance change with different cluster numbers? To this end, we perform experiments on Cifar-10 to evaluate the stabilities of different methods w.r.t. number of clusters. Fig. \ref{fig_4} illustrates the performance changes by varying the cluster numbers from $5$ to $40$ with an interval of $5$. Interestingly, the performance (\textit{i.e.}, ACC, NMI and F-score) of HSIC-based methods increases when the cluster number increases from $5$ to $10$, but then sharply drops using more than $10$ clusters. This suggests that $10$ is the optimal number of clusters. Notably, `purity' can not trade off the precise clustering evaluation against the number of clusters \cite{manning2008}. Importantly, the clustering performance of HSIC in most cases is better than all the compared methods, and HSIC-based methods hold the first three best results. This shows that HSIC is adaptive to different cluster numbers and can be potentially used to predict the `optimal' number of clusters.

\section{Conclusion}\label{conc}
In this paper, we proposed a highly-economized multi-view clustering framework, dubbed HSIC, to jointly learn the compressive binary representations and robust discrete cluster structures. Specifically, HSIC collaboratively integrated the heterogeneous features into the common binary codes, where the sharable and individual information of multiple views were exploited. Meanwhile, a robust cluster structure learning model was developed to improve the clustering performance. Moreover, an effective alternating optimization algorithm was introduced to guarantee the high-quality discrete solutions. Extensive experiments on large-scale multi-view datasets demonstrate the superiority of HSIC over the state-of-the-art methods in terms of clustering performance with significantly reduced computational time and memory footprint.

\bibliographystyle{splncs04}
\bibliography{egbib}

\end{document}